\documentstyle{article}
\def\report{544-97}		
\title{Hierarchical Non-Emitting Markov Models%
\thanks{Both authors are partially supported by Young Investigator
Award IRI-9258517 to the first author from the National Science
Foundation.  Our implementation of the statistical language models
described herein used the library of practical abstractions
\cite{ristad-yianilos:libpa}.  An abbreviated version of this report
was presented at the 1997 Machines that Learn workshop (Snowbird,
April 1--4) and at the 35th Annual Meeting of the ACL (Madrid, July
7--11) \cite{ristad-thomas:acl97}.  Thomas Poetter helped correct some
typographic errors.}}
\author{Eric Sven Ristad \and Robert G. Thomas}
\date{May 1997; Revised January 1998}

\def\f#1{\fbox{#1}}
\def\ith{$i^{\rm th}$}
\def\tuple<#1>{\langle #1 \rangle}
\newtheorem{theorem}{Theorem}
\newtheorem{lemma}{Lemma}[section]
\newcommand{\proof}{\noindent{\bf Proof. }}
\def\qed{\large ~$\Box$ \normalsize}

\begin{document}

\makeatletter
\def\abstract{ \vfil
\begin{center}{\bf Abstract}\end{center}}
\begin{titlepage}
\let\footnotesize\small \let\footnoterule\relax
\leavevmode
\vskip .7in
\begin{center}
{\Large\bf \@title \par}
\vskip 1pc
\begin{tabular}[t]{c}\@author 
\end{tabular}
 
\vskip 1pc
Department of Computer Science\par
Princeton University\par
Research Report CS-TR-\report\par
\@date

\end{center}
\noindent
\@thanks

\begin{abstract}
We describe a simple variant of the interpolated Markov model with
non-emitting state transitions and prove that it is strictly more
powerful than any Markov model.  More importantly, the non-emitting
model outperforms the classic interpolated model on natural
language texts under a wide range of experimental conditions, with
only a modest increase in computational requirements.  The
non-emitting model is also much less prone to overfitting.
\end{abstract}
\vskip 1pc

\end{titlepage}

 \section{Introduction}

The Markov model has long been the core technology of statistical
language modeling.  Many other models have been proposed, but none has
offered a better combination of predictive performance, computational
efficiency, and ease of implementation.  Here we add hierarchical
non-emitting state transitions to the Markov model.  Although the
states in our model remain Markovian, the model itself is no longer
Markovian because it can represent unbounded dependencies in the state
order distribution.  Consequently, the non-emitting Markov model is
strictly more powerful than any Markov model, including the context
model \cite{rissanen:83b,rissanen:86b,willems-etal:95}, the backoff
model \cite{cleary-witten:84,katz:87}, and the interpolated Markov
model \cite{jelinek-mercer:80,mackay-peto:94}.  More importantly, the
non-emitting model consistently outperforms the best Markov models on
natural language texts, under a wide range of experimental conditions.
The non-emitting model is also nearly as computationally efficient and
easy to implement as the interpolated Markov model.

The remainder of our report consists of five sections and one
appendix.  In section~\ref{background-section}, motivate the
fundamental problem of time series prediction, which is to combine the
probabilities of events of different orders.
Section~\ref{interpolate-section} reviews the interpolated Markov
model and briefly demonstrates the equivalence of interpolated models
and basic Markov models of the same model order.  Next, we introduce
the hierarchical non-emitting Markov model in
section~\ref{model-section}, and prove that even a second order
non-emitting model is strictly more powerful than any Markov model, of
any model order.  Section~\ref{estimation-section} provides efficient
algorithms to optimize the parameters of a non-emitting model on data.
In section~\ref{empirical-section}, we report empirical results for
the interpolated model and the non-emitting model on the Brown corpus
and Wall Street Journal.  Finally, in section~\ref{conclude-section}
we conjecture that the non-emitting model excels empirically because
it imposes a pseudo-Bayesian discipline on maximum likelihood
techniques.  Appendix~\ref{backoff-appendix} reviews the backoff model
and explains how to construct a non-emitting backoff model that is
strictly more powerful than any backoff model.

Our notation is as follows.  Let $A$ be a finite alphabet of distinct
symbols, $|A| = k$, and let $x^T \in A^T$ denote an
arbitrary string of length $T$ over the alphabet $A$.  Then $x_i^j$
denotes the substring of $x^T$ that begins at position $i$ and ends at
position $j$.  For convenience, we abbreviate the unit length
substring $x_i^i$ as $x_i$ and the length $t$ prefix of $x^T$ as
$x^t$.

 \section{Time Series Prediction}\label{background-section}

A time series model must assign accurate probabilities to strings of
unbounded length.  Yet unbounded strings don't occur in recorded
histories, which are always finite.  Therefore, to estimate the
probabilities of unbounded strings from a finite corpus, we must
assume that each symbol in a given string depends only on a finite
number of (equivalence classes of) contexts.  The most widely adopted
independence assumption is the order $n$ Markov assumption, which
states that each symbol depends only on the immediately preceding $n$
symbols, and is conditionally independent of the distant past.
\begin{displaymath}
\begin{array}{lcl}
p(x^T|T) & = & \prod_{t=1}^T p(x_t|x^{t-1}) \\
	 & \approx & \prod_{t=1}^T p(x_t|x_{t-n}^{t-1})
\end{array}
\end{displaymath}

The simplest statistical model to incorporate an order $n$ Markov
assumption is the basic Markov model.  A basic Markov model $\phi =
\tuple<A,n,\delta_n>$ consists of an alphabet $A$, a model order $n$,
$n\geq 0$, and the state transition probabilities $\delta_n:A^n\times
A\rightarrow [0,1]$.  With probability $\delta_n(y|x^n)$, a Markov
model in the state $x^n$ will emit the symbol $y$ and transition to
the state $x_2^ny$.  Therefore, the probability
$p_m(x_t|x^{t-1},\phi)$ assigned by an order $n$ basic Markov model
$\phi$ to a symbol $x^t$ in the history $x^{t-1}$ depends only on the
last $n$ symbols of the history.
\begin{equation}\label{basic-evaluate}
  p_m(x_t|x^{t-1},\phi) = \delta_n(x_t|x_{t-n}^{t-1})
\end{equation}
Since the Markov model contains only a finite number of parameters,
it is in principle possible to estimate their values directly from
data.  All that remains is to choose the model order.

In real-world time series problems, the future depends on the entire
past, even if only weakly.  In order to more closely approximate a
real-world source, we would like our model order to be as large as
possible.  Yet we have only a finite amount of training data from
which to estimate our model parameters.  An order $n$ Markov model
over an alphabet of $k$ symbols has $k^{n+1}$ events, while a corpus
of length $T$ has at most $T-n$ distinct events of order $n$.  The
exponential growth in events quickly exceeds the size of all available
training data, and nearly all the higher-order events do not occur in
the training data.

This tension between model complexity and data sparsity is fundamental
to time series modeling.  The probabilities of the lower order events
can be more accurately estimated from the available training data,
while the higher order events are better able to model complex
real-world sources.  An effective model, then, must include individual
events of both higher and lower orders.

The two most widely-used techniques for combining individual events of
varying orders are backoff and interpolation.  In an interpolated
model, the transition probabilities from lower and higher order states
are combined stochastically using mixing parameters.  In a backoff
model, the event probabilities are combined according to a partial
order which typically favors higher order events over lower order
events.  In section~\ref{interpolate-section} and
appendix~\ref{backoff-appendix}, we show that backoff models and
interpolated models are formally equivalent to basic Markov models.
Therefore, backoff and interpolation are simply parameter estimation
schemes for basic Markov models.

 \section{Interpolation}\label{interpolate-section}

Here we introduce the interpolated Markov model and explain why the
interpolated model class is equivalent to the class of basic Markov
models.  In the next section~\ref{model-section}, we introduce
hierarchical non-emitting state transitions to the Markov model, and
prove that the new non-emitting models are no longer Markovian even
though their states are.

In the interpolated Markov model, the transition probabilities from
states of different orders are combined using state-conditional mixing
parameters.  The mixing parameters smooth the transition probabilities
from higher order states with those from lower order states
\cite{jelinek-mercer:80}.  Mixing the transition probabilities from
states of different orders results in more accurate predictions than
can be obtained from any fixed model order.

Formally, an interpolated Markov model $\phi =
\tuple<A,n,\delta,\lambda>$ consists of a finite alphabet $A$, a
maximal model order $n$, the state transition probabilities
$\delta=\delta_0\ldots\delta_n$, $\delta_i: A^i\times A \rightarrow
[0,1]$, and the state-conditional interpolation parameters $\lambda:
A^n\times [0,n]\rightarrow [0,1]$.  The state order is a hidden
variable.  The probability assigned by an interpolated model is a
linear combination of the probabilities assigned by all the lower
order Markov models.
\begin{equation}\label{interpolated-evaluate}
  p_c(y|x^n,\phi) = \sum_{i=0}^n \delta_i(y|x^i) \lambda(i|x^n)
\end{equation}

An interpolated model is a valid probability model if every
$\delta_i(\cdot|x^i)$ and every $\lambda(i|x^n)$ is valid.  It is
nonzero for all strings $A^*$ if $\delta_0(\cdot)$ is strictly
positive for all symbols $A$ and no $\lambda(i|x^n)$ is unity when
$\delta(\cdot|x^i)$ is zero for some symbol.

Estimating the $O(nk^n)$ state interpolation probabilities is
considerably easier than estimating the $O(k^{n+1})$ state transition
probabilities in an order $n$ Markov model.  To begin with, we set
$\lambda(i|x^n)$ to 0 if the order $i$ state $x^i$ is novel.  Now we
need only to estimate the $O(nT)$ interpolation parameters that have
been observed in the training data.

Nonetheless, there are still too many interpolation parameters to be
accurately estimated.  Further refinements are necessary to improve
predictive performance.  One refinement is to group similar parameters
into equivalence classes and then constrain them to take the same
values.  This is called parameter tying.  At one extreme, each
state-conditional interpolation distribution is its own equivalence
class.  At the other extreme, all interpolation probabilities are tied
together and we have the state-independent interpolated Markov model
\begin{equation}\label{tied-interpolated-evaluate}
  p_c(y|x^n,\phi) = \sum_{i=0}^n \delta_i(y|x^i) \lambda_i
\end{equation}
with only $n+1$ interpolation parameters.  While parameter tying can
improve performance, reducing state-conditional interpolation to
state-independent interpolation results in poor performance.

A hierarchical parameterization of the full state-conditional
interpolation is more effective.  Let $\lambda_i:A^i \rightarrow
[0,1]$ be the set of \ith\ order state interpolation parameters, where
$\lambda_i(x^i)$ is the probability of using the \ith order state
transition probability $\delta_i(\cdot|x^i)$, conditioned on the
decision not to use any higher order state transition probability.
\begin{displaymath}
  \lambda(i|x^n) = \lambda_i(x_{n+1-i}^n) \prod_{j=i+1}^{n} (1-\lambda_j(x_{n+1-j}^n))
\end{displaymath}
Then the probability $p_c(y|x^n,\phi)$ that the state $x^n$ will emit
the symbol $y$ has a particularly simple form
\begin{equation}\label{hierarchy-interpolated-evaluate}
\begin{array}{cll}
 p_c(y|x^i,\phi)
	& = & \lambda_i(x^i)\delta_i(y|x^i) \\
	&   & + (1-\lambda_i(x^i)) p_c(y|x_2^i, \phi)
\end{array}
\end{equation}
where $\lambda_i(x^i) = 0$ for $i\geq n$, and therefore
$p_c(x_t|x^{t-1},\phi) = p_c(x_t|x_{t-n}^{t-1},\phi)$, ie., the
prediction depends only on the last $n$ symbols of the history.

A quick glance at the form of (\ref{interpolated-evaluate}) and
(\ref{basic-evaluate}) reveals the fundamental simplicity of the
interpolated Markov model.  Every interpolated model is equivalent to
a basic Markov model of the same order, and every basic Markov model
is an interpolated model of the same order.  We may convert an
interpolated model $\phi$ into a basic model $\phi^\prime$ of the same
model order $n$, simply by setting $\delta_n^\prime(y|x^n)$ equal to
$p_c(y|x^n,\phi)$ for all states $x^n\in A^n$ and symbols $y\in A$.
Thus, the class interpolated Markov models is extensionally equivalent
to the class of basic Markov models.

 \section{Non-Emitting Transitions}\label{model-section}

In the previous section, we explained how to combine events of varying
orders using interpolation and backoff.  Interpolation and backoff
both use the probabilities of lower events to estimate the
probabilities of higher order events.  As a result, interpolated and
backoff models are extensionally equivalent to each other and to basic
Markov models of the same order.  In this section, we explain how to
combine events of varying orders using non-emitting state transitions.

The central idea is to allow actual non-emitting transitions between
events of different orders.  Unlike interpolation and backoff,
non-emitting transitions are not merely an estimation method -- they
actually increase the expressive power of the model class.  As a
result, non-emitting models are strictly more powerful than the class
of basic Markov models.  The next section~\ref{estimation-section}
provides efficient algorithms to evaluate the probability of a string
according to a non-emitting model and to optimize the parameters of a
non-emitting model on data.

A non-emitting mixture Markov model $\phi =
\tuple<A,n,\delta,\lambda>$ consists of a finite alphabet $A$, a
maximal model order $n$, the emitting state transition probabilities
$\delta_i: A^i\times A \rightarrow [0,1]$, and the non-emitting state
transition probabilities $\lambda_i: A^i\times [0,n]\rightarrow
[0,1]$.  The non-emitting model alternates between non-emitting and
emitting transitions according to the $\lambda$ and $\delta$
parameters, respectively.  The parameter $\lambda(j|x^i)$ specifies
the probability that the model will transition from the state $x^i$ to
the state $x^j$ without emitting a symbol.  The parameter
$\delta_j(y|x^j)$ specifies the probability that the model will emit
the symbol $y$ from the state $x^j$ and transition to the successor
state $x^jy$.  Then the probability $p_\epsilon(y^j|x^i,\phi)$
assigned to a string $y^j$ in the state $x^i$ has the form
\begin{equation}\label{nonemit-evaluate}
 p_\epsilon(y^j|x^i,\phi) =
   \sum_{l=0}^i  \lambda(l|x^i) \delta_l(y_1|x^l) p_\epsilon(y_2^j|x^ly,\phi)
.
\end{equation}

When the model order is sufficiently high, then a hierarchical
parameterization of the non-emitting transition probabilities may
improve performance.  With probability $1-\lambda_i(x^i)$, a
hierarchical non-emitting model will transition from the state $x^i$
to the state $x_2^i$ without emitting a symbol.  With probability
$\lambda_i(x^i)\delta_i(y|x^i)$, the model will transition from the
state $x^i$ to the state $x^iy$ and emit the symbol $y$.

Therefore, the probability $p_\epsilon(y^j|x^i,\phi)$ assigned to a
string $y^j$ in the history $x^i$ by a hierarchical non-emitting model
$\phi$ has the recursive form (\ref{hierarchy-nonemit-evaluate}),
\begin{equation}\label{hierarchy-nonemit-evaluate}
\begin{array}{lcl}
p_\epsilon(y^j|x^i,\phi) 
	& = & \lambda_i(x^i)\delta_i(y_1|x^i) p_\epsilon(y_2^j|x^i y_1,\phi) \\
	&   & + (1-\lambda_i(x^i)) p_\epsilon(y^j|x_2^i, \phi)
\end{array}
\end{equation}
where $\lambda_i(x^i) = 0$ for $i > n$ and $\lambda_0(\epsilon) = 1$.
Note that, unlike the basic Markov model,
$p_\epsilon(x_t|x^{t-1},\phi) \neq p_\epsilon(x_t|x_{t-n}^{t-1},\phi)$
because the state order distribution of the non-emitting model depends
on the prefix $x^{i-n}$.  This simple fact will allow us to establish
that there exists a non-emitting model that is not equivalent to any
Markov model.

Lemma~\ref{state1-lemma} states that there exists a non-emitting model
$\phi$ that cannot be converted into an equivalent basic model of any
order.  There will always be a string $x^T$ that distinguishes the
non-emitting model $\phi$ from any given basic model $\phi^\prime$
because the non-emitting model can encode unbounded dependencies in
its state distribution.
\begin{lemma}\label{state1-lemma}
$\exists \phi \
	\forall \phi^\prime \
		\exists x^T \in A^* \
			[ p_\epsilon(x^T|\phi, T) \neq p_m(x^T|\phi^\prime, T) ]$
\end{lemma}
\proof 
The idea of the proof is that our non-emitting model will encode the
first symbol $x_1$ of the string $x^T$ in its state distribution, for
an unbounded distance.  This will allow it to predict the last symbol
$x_T$ using its knowledge of the first symbol $x_1$.  The basic 
model will only be able predict the last symbol $x_T$ using the
preceding $n$ symbols, and therefore when $T$ is greater than $n$, we
can arrange for $p_\epsilon(x^T|\phi, T)$ to differ from any
$p_m(x^T|\phi^\prime, T)$, simply by our choice of $x_1$.

The smallest non-emitting model capable of exhibiting the required
behavior has order 2.  Lower order non-emitting models are equivalent
to interpolated models of the same order, with the same parameters.
The non-emitting transition probabilities $\lambda$ and the interior
of the string $x_2^{T-1}$ will be chosen so that the non-emitting
model is either in an order 2 state or an order 0 state, with no way
to transition from one to the other.  The first symbol $x_1$ will
determine whether the non-emitting model goes to the order 2 state or
stays in the order 0 state.  No matter what probability the basic
model assigns to the final symbol $x_T$, the non-emitting model can
assign a different probability by the appropriate choice of $x_1$,
$\delta_0(x_T)$, and $\delta_2(x_T|x_{T-2}^{T-1})$.

Consider the second order non-emitting model over a binary alphabet
with $\lambda(0)=1$, $\lambda(1)=0$, and $\lambda(11)=1$ on strings in
$A 1^* A$.  When $x_1=0$, then $x_2$ will be predicted using the 1st
order model $\delta_1(x_2|x_1)$, and all subsequent $x_t$ will be
predicted by the second order model $\delta_2(x_t|x_{t-2}^{t-1})$.
When $x_1=1$, then all subsequent $x_t$ will be predicted by the zeroth
order model $\delta_0(x_t)$.  Thus for all $t>p$,
$p_\epsilon(x_t|x^{t-1}) \neq p_\epsilon(x_t|x_{t-p}^{t-1})$ for any
fixed $p$, and no basic model is equivalent to this simple
non-emitting model.  
\qed

Every basic model is a non-emitting model, with the appropriate choice
of non-emitting transition probabilities.
\begin{lemma}\label{state2-lemma}
$\forall \phi \
	\exists \phi^\prime \
		\forall x^T \in A^* \
			[ p_\epsilon(x^T|\phi^\prime, T) = p_m(x^T|\phi, T) ]$
\end{lemma}
\proof
A basic model $\phi = \tuple<A,n,\delta_n>$ is equivalent to a
non-emitting model $\phi^\prime =
\tuple<A,n,\delta^\prime,\lambda^\prime>$ where $\delta_n^\prime =
\delta_n$ and $\lambda^\prime(n|x^n) = 1$ for all $x^n$.  In the
hierarchical parameterization, $\lambda^\prime(x^n) = 1$ for all
$x^n$.
\qed

Therefore, the class ${\cal P}_\epsilon$ of non-emitting Markov
distributions is strictly more powerful than the class ${\cal P}_m$ of
basic Markov distributions.
\begin{theorem}
${\cal P}_m \subset {\cal P}_\epsilon$
\end{theorem}
\proof 
${\cal P}_m \neq {\cal P}_\epsilon$ by lemma~\ref{state1-lemma} 
and ${\cal P}_m \subseteq {\cal P}_\epsilon$ by lemma~\ref{state2-lemma}.
\qed

Since interpolated models and backoff models are equivalent to basic
Markov models, we have as a corollary that non-emitting Markov models
are strictly more powerful than interpolated and backoff models.  Note
that non-emitting Markov models are considerably less powerful than
the full class of stochastic finite state automata because
their states are Markovian.  For the same reason, non-emitting models
are also less powerful than the full class of hidden Markov models.

Let us now turn to the algorithms required to evaluate the probability
of a string according to a non-emitting mixture model and to optimize
the non-emitting state transitions on a training corpus.

 \section{Estimation}\label{estimation-section}

Here we present an efficient expectation-maximization (EM) algorithm
to optimize the parameters of a hierarchical non-emitting mixture
model on data.  An EM algorithm iteratively maximizes the probability
of the training data according to the model by computing the
expectation of model parameters on the data and then updating the
model parameters to maximize those expectations
\cite{baum-eagon:67,baum-etal:70,dempster-etal:77}.

The non-emitting mixture model is sufficiently expressive that any
maximum likelihood estimator will overfit its parameters to the training
corpus.  Unseen events will be assigned zero probability, and the
overfit model will fail to accurately predict the future.  The
traditional solution to this problem for interpolated Markov models is
cross-estimation \cite{jelinek-mercer:80}.  Cross-estimation
repeatedly partitions the training data into two blocks and optimizes
the mixing parameters on one block after initializing the state
transition parameters on the other block.  We present a traditional
cross-estimation algorithm for hierarchical non-emitting models.

We begin by partitioning the training corpus into a fixed set of
blocks $\bf B$.  Ideally our partition is linguistically meaningful
and roughly uniform, but neither condition is essential.  For example,
we might divide a natural language text corpus on sentence, paragraph,
or article boundaries.  Next we call {\sc cross-estimate\/}(${\bf
B}$,$\phi$) on our hierarchical non-emitting model $\phi$.

{\sf
\begin{tabbing}
aaa \= aaa \= aaa \= aaa \= \kill
{\sc cross-estimate}(${\bf B}$,$\phi$) \\
1. \> Until convergence\\
2. \> \> Initialize $\lambda^+,\lambda^-$ to zero;\\
3. \> \> For each block $B_i$ in ${\bf B}$ \\
4. \> \> \> Initialize $\delta$ using ${\bf B} - B_i$; \\
5. \> \> \> {\sc expectation-step\/}($B_i$,$\phi$,$\lambda^+$,$\lambda^-$); \\
6. \> \> {\sc maximization-step\/}($\phi$,$\lambda^+$,$\lambda^-$); \\
7. \> Initialize $\delta$ using ${\bf B}$;
\end{tabbing}
}

The variables $\lambda^+(x^i)$ and $\lambda^-(x^i)$ accumulate
expectations for the non-emitting state transition parameter
$\lambda(x^i)$.  $\lambda^+(x^i)$ contains the expectation of emitting
a symbol in state $x^i$, conditioned on being in state $x^i$, while
$\lambda^-(x^i)$ contains the expectation of transitioning to $x_2^i$
without emitting a symbol, conditioned on being in state $x^i$.  Lines
3-5 enumerate all one-block partitions of the training corpus.  The
emitting state transitions $\delta$ are initialized to their maximum
likelihood estimates on the larger block ${\bf B} - B_i$ and then the
non-emitting state transitions $\lambda$ are optimized on the smaller
``withheld'' block $B_i$.

The heart of the algorithm is the {\sc expectation-step\/}()
procedure, which calculates the expectation of the non-emitting
transitions on the string $x^b$ and then increments the
$\lambda^+,\lambda^-$ accumulators.

{\sf
\begin{tabbing}
aaa \= aaa \= aaa \= aaa \= \kill
{\sc expectation-step}($x^b$,$\phi$,$\lambda^+$,$\lambda^-$) \\
1. \> $\alpha$ = {\sc forward}($x^b$,$\phi$); \\
2. \> $\beta$ = {\sc backward}($x^b$,$\phi$); \\
3. \> for $t = b$ downto $1$ \\
4. \> \> for $i = 1$ upto $\min(n,t)$ \\
5. \> \> \> $\lambda^-_t(i) = \alpha_t(i)(1-\lambda_t(i))\beta_t(i-1)$; \\
6. \> \> \> $\lambda^+_{t-1}(i-1) +=
		\alpha_{t-1}(i-1)\lambda_{t-1}(i-1)\delta_{t-1}(i-1)\beta_t(i)$; \\
7. \> \> if ($t > n$)
	[ $\lambda^+_{t-1}(n) += 
		\alpha_{t-1}(n)\lambda_{t-1}(n)\delta_{t-1}(n)\beta_t(n)$; ]
\end{tabbing}
}

The forward variable $\alpha_t(i)$ contains the probability $p(x^t,
o_t = i | \phi)$ that the model $\phi$ generated the prefix $x^t$ and
terminated in the order $i$ state.  The backward variable $\beta_t(i)$
contains the probability $p(x_{t+1}bT | x^t, o_t = i, \phi)$ that the
model $\phi$ generated the suffix $x_{t+1}^b$ given that it was in the
order $i$ state at time $t$.  To simplify the notation, we define
$\lambda_t(i)$ to be the probability $\lambda(x_{t+1-i}^t)$ of emitting
a symbol from the \ith\ order state at time $t$, given that we are in
that state.  We also define $\delta_t(i)$ to be the probability
$\delta_i(x_{t+1}|x_{t+1-i}^t)$ of the emitting transition from state
$x_{t+1-i}^t$ to state $x_{t+1-i}^{t+1}$.

The {\sc expectation-step\/}() algorithm requires $O(nb)$ time and
space for an order $n$ non-emitting model on a string $x^b$ of length
$b$.  A comparable interpolated model can take an expectation step in
$O(nb)$ time and $O(1)$ space \cite{bahl-etal:91}.  While the
difference between $O(nb)$ and $O(1)$ space can be considerable, the
additional space requirements of the non-emitting algorithm are small
when compared to the cost of storing all the model parameters.  An
order $n$ mixture model has $O(nT)$ parameters for a training corpus
of size $T$, and the training corpus is typically an order of
magnitude larger than the withheld block.

{\sf
\begin{tabbing}
aaa \= aaa \= aaa \= aaa \= \kill
{\sc forward}($x^T$,$\phi$) \\
1. \> $\alpha_0(0) = 1$; \\
2. \> for $t = 1$ upto $T-1$ \\
3. \> \> for $i = \min(n-1, t)$ downto $0$ \\
4. \> \> \> $\alpha_t(i) += \alpha_t(i+1) (1-\lambda_t(i+1))$; \\
5. \> \> \> $\alpha_{t+1}(i+1) := \alpha_t(i)\lambda_t(i)\delta_t(i)$; \\
6. \> \> if ($t\geq n$) [ $\alpha_{t+1}(n) += \alpha_t(n)\lambda_t(n)\delta_t(n)$; ] \\
7. \> return($\alpha$);
\end{tabbing}
}

{\sf
\begin{tabbing}
aaa \= aaa \= aaa \= aaa \= \kill
{\sc backward}($x^T$,$\phi$) \\
1. \> for $i = 0$ upto $\min(n-1,T-1)$; \\
2. \> \> $\beta_{T-1}(i) = \lambda_{T-1}(i)\delta_{T-1}(i)$; \\
3. \> if ($T > n$) [ $\beta_{T-1}(n) = \lambda_{T-1}(n)\delta_{T-1}(n)$; ] \\
4. \> for $t = T-1$ downto $1$ \\
5. \> \> for $i = 1$ upto $\min(n,t)$ \\
6. \> \> \> $\beta_t(i) += (1-\lambda_t(i))\beta_t(i-1)$; \\
7. \> \> \> $\beta_{t-1}(i-1) = \lambda_{t-1}(i-1)\delta_{t-1}(i-1)\beta_t(i)$; \\
8. \> \> if ($t > n$) [ $\beta_{t-1}(n) = \lambda_{t-1}(n)\delta_{t-1}(n)\beta_T(n)$; ] \\
9. \> return($\beta$);
\end{tabbing}
}

The {\sc forward\/}() and {\sc backward}() algorithms each require
$O(nT)$ time and space.  It is possible to evaluate the probability
$p_\epsilon(x^T|\phi)$ of a string $x^T$ according to an order $n$
non-emitting model $\phi$ in $O(nT)$ time and $O(n)$ space.  In
contrast, it is possible to evaluate the probability $p_c(x^T|\phi)$
according to an interpolated model in $O(nT)$ time and $O(1)$ space.
Again, the small additional cost in space is negligible when compared
to the cost of storing the model parameters.

Having done all the work in the expectation step, the maximization
step is straightforward.  

{\sf
\begin{tabbing}
aaa \= aaa \= aaa \= aaa \= \kill
{\sc maximization-step}($\phi$,$\lambda^+$,$\lambda^-$) \\
1. \> Forall states $x^i$ in $A^{\leq n}$ \\
2. \> \> $\bar{\lambda(x^i)} :=  \lambda^+(x^i) / (\lambda^+(x^i) + \lambda^-(x^i))$; \\
\end{tabbing}
}

Line 2 reestimates each non-emitting state transition parameter
$\lambda(x^i)$ as the expectation of emitting a symbol from that state
divided by the expectation of being in that state.  In order to ensure
that no non-emitting state transition parameter $\lambda(x^i)$ is ever
reestimated to 0 or 1, we typically initialize each accumulator to a
small positive number (eg., 0.1) instead of zero.

When $\lambda$ parameters are tied, then their $\lambda^+$ and
$\lambda^-$ expectations must be pooled before they are updated.  Let
$\tau(x^i)$ be the equivalence class of $x^i$ under the tying scheme
$\tau$.  For simplicity, imagine $\tau(x^i)$ to be an index.  All
algorithms in this section would use the tied parameter
$\lambda(\tau(x^i))$ instead of the untied parameter $\lambda(x^i)$.
The {\sc tied-expectation-step\/}() algorithm would increment the
$\lambda^+(\tau(x^i))$ and $\lambda^-(\tau(x^i))$ accumulators, and
the {\sc tied-maximization-step\/}() algorithm would be as follows.

{\sf
\begin{tabbing}
aaa \= aaa \= aaa \= aaa \= \kill
{\sc tied-maximization-step}($\phi$,$\lambda^+$,$\lambda^-$,$\pi$) \\
1. \> Forall classes $i$ in $\tau(A^{\leq n})$ \\
2. \> \> $\bar{\lambda(i)} := \lambda^+(i) / (\lambda^+(i) + \lambda^-(i))$; \\
\end{tabbing}
}

In some situations, cross-estimation may be approximated by
forward-estimation.  Like cross-estimation, forward-estimation
initializes the $\delta$ parameters on one text block and optimizes
the $\lambda$ parameters on another block.  Forward-estimation uses
only a single text partition whereas cross-estimation uses all
one-block text partitions.  As result, forward-estimation is
considerably faster than cross-estimation, both in the amount of time
required per iteration and in the number of iterations until
convergence.  Unfortunately, it can lead to inferior results when
there are too many mixing parameters.

{\sf
\begin{tabbing}
aaa \= aaa \= aaa \= aaa \= \kill
{\sc forward-estimate}($B_\delta$,$B_\lambda$,$\phi$) \\
1. \> Until convergence\\
2. \> \> Initialize $\lambda^+,\lambda^-$ to zero;\\
3. \> \> Initialize $\delta$ using $B_\delta$; \\
4. \> \> {\sc expectation-step\/}($B_\lambda$,$\phi$,$\lambda^+$,$\lambda^-$); \\
5. \> \> {\sc maximization-step\/}($\phi$,$\lambda^+$,$\lambda^-$); \\
6. \> Initialize $\delta$ using $B_\delta\cup B_\lambda$;
\end{tabbing}
}

 \paragraph{Implementation Note.}

Unless the corpus and the alphabet size are very small, then the
$\alpha_t(i)$ and $\beta_t(i)$ values used in the {\sc
expectation-step\/}() procedure will exceed the representational range
of double precision IEEE floating point numbers.  When this happens, a
floating point exception will occur and an alternate representation
must be used for the probability values.  The simplest approach is to
use a logarithmic representation.  Multiplication and division of
probability values is straightforward in a logarithmic representation.
\begin{displaymath}\begin{array}{lcl}
 \log(x\cdot y)	& = & \log(x)+\log(y) \\
 \log(x / y)	& = & \log(x)-\log(y)
\end{array}\end{displaymath}
Addition of logarithmic probability values is more costly, and care
must be taken to avoid underflow.
\begin{displaymath}
\log(x + y) = 
 \left\{\begin{array}{ll}
  \log(x)				  & \mbox{ if } \log(y)-\log(x)\leq\Lambda \\
  \log(x) + \log(1+\exp(\log(y)-\log(x))) & \mbox{ otherwise }
 \end{array}\right.
\end{displaymath}
Here $\Lambda$ is the smallest representable exponent, for example,
-707.7 for IEEE double precision floating point numbers when the
logarithms are natural (ie., base $e$).  This test is necessary to
avoid underflow in the call to $\exp()$.  

While it is simple to implement, logarithmic arithmetic can be 15-50
times slower than straight probability arithmetic, depending on the
speed of the floating point unit and the math library provided with
the operating system.  For this reason, our implementation used an
extended exponent representation from the library of practical
abstractions \cite{ristad-yianilos:libpa}.  This \verb|balanced_t|
module provides single precision floating point numbers with 32 bit
exponents.  It is 1.5 to 3.0 times faster than the logarithmic
representation, depending on the machine.  

When computation time is at a premium, then the most effective
solution is to periodically scale the probability values in the
$\alpha_t(i)$ and $\beta_t(i)$ arrays to keep them in an acceptable
range.  Scaling is more difficult to implement than logarithmic
arithmetic or \verb|balanced_t| arithmetic, and it is inherently
nonmodular.

 \section{Empirical Results}\label{empirical-section}

The ultimate measure of a statistical model is its predictive
performance in the domain of interest.  To take the true measure of
non-emitting models for natural language texts, we evaluate their
performance as character models on the Brown corpus
\cite{francis-kucera:82} and as word models on the Wall Street
Journal.  Our results show that the non-emitting Markov model
consistently gives better predictions than the traditional
interpolated Markov model under equivalent experimental conditions.
In all cases we compare non-emitting and interpolated models of
identical model orders, with the same number of parameters.  Note that
the non-emitting bigram and the interpolated bigram are equivalent.

\begin{center}
\begin{tabular}{l|rrr}
{\it Corpus}&  {\it Alphabet} & {\it Size} & {\it Blocks} \\ \hline
Brown 	    &     90 &  6,004,032 &  21 \\
WSJ 1989    & 20,293 &  6,219,350 &  22 \\
WSJ 1987-89 & 20,092 & 42,373,513 & 152 \\
\end{tabular}
\end{center}

All $\lambda$ values were initialized uniformly to $0.5$ and then
optimized using cross-estimation on the first 90\% of each corpus.
The remaining 10\% percent of each corpus was used to evaluate model
performance.  While this validation paradigm exposes the models to
nonstationarity, it is simple to understand and easily reproduced.

We consider a single parameter tying scheme, in which all states with
the same frequency and diversity are considered equivalent.  The
frequency $c(x^i)$ of a state is the number of times that the string
$x^i$ occurred in the training corpus.  The diversity $q(x^i) \doteq
|\{ y: c(x^iy) > 0 \}|$ of a state is the number of distinct symbols
observed in the state.  Experience with multinomial prediction
suggests that frequency and diversity are necessary to accurately
estimate the likelihood of novel symbols \cite{ristad:pu495-95}.

In related work \cite{thomas:phd}, Thomas compares the performance of
the interpolated and non-emitting models on the Brown corpus and Wall
Street Journal with ten different parameter tying schemes.  His
experiments confirm that some parameter tying schemes improve model
performance, although to a lesser degree when cross-estimation is
used.  The non-emitting model consistently outperformed the
interpolated model on both corpora for all ten parameter tying
schemes.  Thomas shows that our frequency-diversity parameter tying
scheme is one of the more effective parameter schemes.

 \subsection{Brown Corpus}
\nocite{brown-etal:92}

Our first set of experiments were with character models on the Brown
corpus \cite{francis-kucera:82}.  The Brown corpus is an eclectic
collection of English prose, containing 6,004,032 characters
partitioned into 500 files.  We performed 10 iterations of cross
estimation on 21 blocks.  Results are reported as per-character test
message entropies (bits/char), $-\frac{1}{v}\log_2 p(y^v|v)$.  The
non-emitting model outperforms the interpolated model for all
nontrivial model orders, particularly for larger model orders.  The
non-emitting model is considerably less prone to overfitting.  After
10 EM iterations, the untied order 9 non-emitting model scores 1.996
bits/char while the untied order 9 interpolated model scores 2.334
bits/char.  The untied non-emitting model even outperforms the {\it
tied\/} interpolated model for all nontrivial model orders.

\begin{center}
\begin{tabular}{c|cccc}
Model & \multicolumn{2}{c}{Interpolation} & \multicolumn{2}{c}{Non-Emitting} \\
order & untied & tied & untied & tied \\
\hline
1 & \f{3.602} & \f{3.602} & \f{3.602} & \f{3.602} \\
2 & 2.950 & 2.950 & \f{2.946} & \f{2.946} \\
3 & 2.490 & 2.486 & \f{2.473} & \f{2.473} \\
4 & 2.231 & 2.218 & 2.193 & \f{2.192} \\
5 & 2.149 & 2.112 & 2.076 & \f{2.075} \\
6 & 2.164 & 2.082 & 2.031 & \f{2.027} \\
7 & 2.212 & 2.077 & 2.015 & \f{2.008} \\
8 & 2.277 & 2.084 & 2.010 & \f{2.000} \\
9 & 2.334 & 2.093 & 2.009 & \f{1.996}
\end{tabular}
\end{center}

We also compared the performance of our techniques with two new
interpolation schemes recently proposed by Potamianos and Jelinek
\cite{potamianos-jelinek:97}.  Their DI-TD scheme uses hierarchical
state-conditional interpolation $\lambda(x^i)$, variable-width
frequency $\times$ order parameter tying, and ``top-down
optimization'' on one withheld block.  Their DI-BU scheme uses general
state-conditional interpolation $\lambda(j|x^i)$, variable-width
frequency $\times$ order parameter tying, and bottom-up optimization
on one withheld block.  The comparison is performed on a modified
version of the Brown corpus, which they provided to us.  This modified
corpus eliminates the unusual punctuation of the original Brown
corpus, reduces the alphabet size from 90 to 79, and separates
distinct linguistic tokens with single spaces.
\begin{center}
\begin{tabular}{l|ccccc}
{\it Corpus}& {\it Alphabet} & {\it Size} & {\it Train} & {\it Test} & {\it Blocks} \\ \hline
Brown (std) & 90 & 6,004,032 & 5,403,629 & 600,403 & 21  \\
Brown (JHU) & 79 & 6,093,662 & 5,607,270 & 486,392 & 21  
\end{tabular}
\end{center}
Another difference between the Potamianos-Jelinek validation paradigm
and ours lies in how the corpus is partitioned into training and
testing blocks.  In our experiments, the test block was the last 10\%
of the Brown corpus -- the last 428 characters from br-n14.txt plus
all files from br-n15.txt through br-r09.txt inclusive.  In the
Potamianos-Jelinek experiments, the test block consisted of complete
sentences chosen uniformly from the entire (modified) Brown corpus.

To this comparison, we added the original interpolation schemes of
Jelinek and Mercer \cite{jelinek-mercer:80} under 10 iterations of
forward-estimation (DI-FE) and cross-estimation (DI-CE).  Both models
used hierarchical state-conditional interpolation $\lambda(x^i)$ and
straight frequency $\times$ diversity parameter tying.  We also added
the hierarchical non-emitting model with straight frequency $\times$
diversity parameter tying, and 10 iterations of forward-estimation
(NE-FE) and cross-optimization (NE-CE).  The results are summarized in
the following table as mean test message entropies (bits/char).
\begin{center}
\begin{tabular}{c|cccc|cc}
Model & \multicolumn{4}{c|}{Interpolation} & \multicolumn{2}{c}{Non-Emitting} \\
order & DI-TD & DI-BU & DI-FE & DI-CE & NE-FE & NE-CE \\ \hline
1 & \f{3.470} & \f{3.470} & 3.478 & 3.478 & 3.478 & 3.478 \\
2 & 2.851 & \f{2.850} & 2.860 & 2.858 & 2.857 & 2.856 \\
3 & 2.328 & 2.326 & 2.337 & 2.331 & 2.328 & \f{2.324} \\
4 & 2.016 & 2.007 & 2.012 & 2.007 & 1.996 & \f{1.991} \\
5 & 1.894 & 1.878 & 1.872 & 1.867 & 1.849 & \f{1.843} \\
6 & 1.853 & 1.831 & 1.820 & 1.815 & 1.789 & \f{1.782} \\
7 & 1.837 & 1.811 & 1.804 & 1.800 & 1.761 & \f{1.754} \\
8 & 1.828 & 1.801 & 1.800 & 1.796 & 1.746 & \f{1.739} \\
9 & 1.824 & 1.796 & 1.802 & 1.798 & 1.738 & \f{1.730} 
\end{tabular}
\end{center}
The non-emitting model consistently outperforms all interpolation
schemes at all model orders above 2, by a significant margin.  The
original Jelinek-Mercer interpolation scheme also tends to outperform
the two new DI-TD and DI-BU schemes at higher model orders, for both
forward-estimation (DI-FE) and cross-estimation (DI-CE).

Note also that the best order 9 result in the Potamianos-Jelinek
paradigm (1.730 bits/char) is considerably better than the best order
9 result in our validation paradigm (1.996 bits/char).  We believe
this is partially attributable to the reduced alphabet size of the
modified corpus, and principally due to the difference in the two
train-test partitions.  The prediction problem posed by our paradigm
is more difficult because the last 10\% of the Brown files are
appreciably different than the first 90\% of the files.

 \subsection{WSJ 1989}

The second set of experiments was on the 1989 Wall Street Journal
corpus, which contains 6,219,350 words.  Our vocabulary consisted of
the 20,293 words that occurred at least 10 times in the entire WSJ
1989 corpus.  All out-of-vocabulary words were mapped to a unique OOV
symbol.  We performed 10 iterations of cross estimation on 22 blocks.
Following standard practice in the speech recognition community,
results are reported as per-word test message perplexities
$p(y^v|v)^{-\frac{1}{v}}$.  The perplexity represents the effective
alphabet size.  Again, the non-emitting model outperforms the
interpolated model for all nontrivial model orders, even without
parameter tying.

\begin{center}
\begin{tabular}{c|cccc}
Model & \multicolumn{2}{c}{Interpolation} & \multicolumn{2}{c}{Non-Emitting} \\
order & untied & tied & untied & tied \\
\hline
1 & 175.2 & \f{174.9} & 175.2 & \f{174.9} \\
2 & 123.7 & 122.8 & 119.6 & \f{119.0} \\
3 & 121.3 & 119.0 & 111.9 & \f{111.1} \\
4 & 123.0 & 117.2 & 110.6 & \f{109.5} \\
5 & 124.5 & 116.3 & 110.4 & \f{109.0}
\end{tabular}
\end{center}

 \subsection{WSJ 1987-89}

The third set of experiments was on the 1987-89 Wall Street Journal
corpus, which contains 42,373,513 words.  Our vocabulary consisted of
the 20,092 words that occurred at least 63 times in the entire WSJ
1987-89 corpus.  Again, all out-of-vocabulary words were mapped to a
unique OOV symbol.  We performed 10 iterations of cross estimation on
152 blocks.  Results are reported as test message perplexities.  As
with the WSJ 1989 corpus, the non-emitting model outperforms the
interpolated model for all nontrivial model orders, even without
parameter tying.

\begin{center}
\begin{tabular}{c|rrrr}
Model & \multicolumn{2}{c}{Interpolation} & \multicolumn{2}{c}{Non-Emitting} \\
order & untied & tied & untied & tied \\
\hline
1 & \f{150.7} & \f{150.7} & \f{150.7} & \f{150.7} \\
2 &  94.0 &  93.9 &  \f{92.1} &  \f{92.1} \\
3 &  89.2 &  88.6 &  \f{83.2} &  \f{83.2}
\end{tabular}
\end{center}

 \subsection{Posthoc Analysis}

In order to understand the striking empirical advantage of the
non-emitting model over the interpolated model, we conducted the
following experiment.  We induced order 9 interpolated and
non-emitting models from the Brown corpus using forward estimation
with no parameter tying.  This configuration was chosen to maximize
the performance difference between the two models.  The resulting
interpolated model predicts the Brown test corpus with 2.4480
bits/char while the resulting non-emitting model predicts the Brown
test corpus with 2.1536 bits/char.

The following table shows the mean state order occupancy
statistics for the two models on the Brown corpus.  
\begin{center}
\begin{tabular}{c|cc}
{\bf Order} & {\bf Interpolated} & {\bf Non-Emitting} \\ \hline
9 & 0.133 & 0.070 \\
8 & 0.120 & 0.090 \\
7 & 0.131 & 0.127 \\[0.1cm]
6 & 0.147 & 0.170 \\
5 & 0.147 & 0.195 \\
4 & 0.130 & 0.173 \\
3 & 0.095 & 0.108 \\[0.1cm]
2 & 0.058 & 0.047 \\
1 & 0.027 & 0.013 \\
0 & 0.011 & 0.003 \\
\hline
  & {\sf 5.639} & {\sf 5.357}
\end{tabular}
\end{center}
As might be expected, the interpolated model spends more time than the
non-emitting model in the higher order states (orders 7-9).  It is
arguably more surprising, however, that the interpolated model also
spends more time in the lower order states (orders 0-2).

One point where the non-emitting model outperforms the interpolated
model is in predicting the space \verb*| | that follows the string 
\verb*|, but now Keith| in the Brown test corpus.  
Unfortunately, the string \verb*| Keith| does not occur in the
training corpus.  Nonetheless, the non-emitting model assigns 209
times more probability than the interpolated model to the event that a
space will follow the string \verb*| Keith|.  According to the
non-emitting model, a space will follow the string \verb*| Keith| with
probability 0.627.  The interpolated model assigns probability 0.003
to the same event.

The reason is somewhat subtle.  On the training corpus, the string
\verb*|eith| is followed by the letter \verb*|e| with near certainty
(0.9973).  As a result, $\lambda$(\verb*|eith|) approaches unity in
both the interpolated and non-emitting models.  Since the model order
9 is sufficiently high, the interpolated model will use the
\verb*|eith| state whenever it occurs and no higher order state is
preferred (see figure~\ref{interpolate-occupancy-figure}).

The hierarchical non-emitting model has no such freedom (see
figure~\ref{nonemit-occupancy-figure}).  In order to reach the
\verb*|eith| state, it must accurately predict every symbol in the
string \verb*|eith|.  Otherwise, it will be forced to a lower order
state along the way.  The transition to a lower order state occurs
when the non-emitting model attempts to predict the symbol \verb*|t|
from the state \verb*|ei|.  Since \verb*|ei| is rarely followed by
\verb*|t| in the training corpus (.0761), the non-emitting model is
forced into the lower order state \verb*|i|, from which it is able to
predict the symbol \verb*|t| with greater probability (.1172).  As a
result, the non-emitting model is never able to reach the \verb*|eith|
state.  Instead, it must predict the space \verb*| | after 
\verb*| Keith| using the state \verb*|ith|.  This works quite well
because \verb*|ith| is followed by \verb*| | with high probability in
the training corpus (0.6136).

\begin{figure}[htb]
\begin{center}
\begin{tabular}{r|ccccccc} 
9 & 0.549 & 0.446 &       &       &       &       &       \\
8 & 0.275 & 0.223 &       &       &       &       &       \\
7 & 0.137 & 0.049 &       &       &       &       &       \\
6 & 0.037 & 0.082 &       &       &       &       &       \\
5 & 0.001 & 0.038 &       &       &       &       &       \\
4 &       & 0.067 &       &       &       &       & 1.000 \\
3 &       & 0.050 &       & 0.376 &       &       &       \\
2 &       & 0.025 & 0.856 & 0.617 & 0.527 & 0.596 &       \\
1 &       & 0.015 & 0.142 & 0.004 & 0.415 & 0.297 &       \\
0 &       & 0.004 & 0.002 & 0.004 & 0.058 & 0.106 &       \\
\hline
  & \verb*| | & \verb*|K| & \verb*|e| & \verb*|i| & \verb*|t| &  \verb*|h| & \verb*| | \\
\hline

  & 0.998 & 0.000 & 0.273 & 0.006 & 0.093 & 0.226 & 0.003
\end{tabular}
\end{center}
\caption[Interpolated state occupancies.]{State occupancy
probabilities for the order 9 interpolated model on part of the Brown
test corpus (2.4480 bits/char).  The horizontal axis represents the
position in the test string and the vertical access represents the
hidden state order.  The bottom column shows the conditional
probability of the symbol, given the hidden state distribution.  Thus
the interpolated model is in the order 4 state \verb*|eith| with
probability at least .9995 when predicting the final symbol, and it
assigns probability 0.003 to this symbol.}
\label{interpolate-occupancy-figure}
\end{figure}

\begin{figure}[htb]
\begin{center}
\begin{tabular}{r|ccccccc} 
9 & 0.166 & 0.121 &       &       &       &       & \\
8 & 0.289 & 0.369 &       &       &       &       & \\
7 & 0.340 & 0.449 &       &       &       &       & \\
6 & 0.161 & 0.016 &       &       &       &       & \\
5 & 0.035 & 0.017 &       &       &       &       & \\
4 & 0.008 & 0.017 &       &       &       &       & \\
3 &       & 0.008 &       & 0.472 &       &       & 0.734 \\
2 &       & 0.002 & 0.939 & 0.527 & 0.848 & 0.749 & 0.187 \\
1 &       & 0.001 & 0.061 & 0.001 & 0.118 & 0.207 & 0.003 \\
0 &       &       &       &       & 0.033 & 0.044 & 0.075 \\
\hline
  & \verb*| | & \verb*|K| & \verb*|e| & \verb*|i| & \verb*|t| &  \verb*|h| & \verb*| | \\
\hline
  & 0.977 & 0.000 & 0.277 & 0.006 & 0.081 & 0.232 & 0.627
\end{tabular}
\end{center}
\caption[Non-emitting state occupancies.]{State occupancy
probabilities for the order 9 non-emitting model on part of the Brown
test corpus (2.1536 bits/char). The horizontal axis represents the
position in the test string and the vertical access represents the
hidden state order.  The bottom column shows the conditional
probability of the symbol, given the hidden state distribution.  Thus
the non-emitting model is in the order 3 state \verb*|ith| with
probability 0.734 when predicting the final symbol, and it assigns
probability 0.627 to this symbol.}
\label{nonemit-occupancy-figure}
\end{figure}

\clearpage
 \subsection{Posterior Tying}

This posthoc analysis led John Lafferty (personal communication) to
suggest that the interpolated model might be able to approximate the
empirical performance of the non-emitting model with a suitable
parameter tying scheme.  According to the non-emitting model, two
states should be considered equivalent if they are equally effective
at predicting the future {\it and\/} they are equally well predicted
by the model.  A state is well-predicted if the string that it
represents is assigned high probability, relative to the other states
available at the time.  A state provides strong predictions if the
entropy of its emitting state transition probabilities is low.

The most effective way for the interpolated model to mimic the
non-emitting model is to tie its states based on their expectations in
the corresponding non-emitting model.  In order to avoid implementing
the non-emitting model, we may reasonably impose a uniform
distribution on the non-emitting state transitions.  And in order to
avoid running the full {\sc expectation-step\/}() algorithm, we may
approximate the non-emitting state expectations by their forward
expectations in $O(nT)$ time and $O(n)$ space.

A further simplification is to use the mean empirical posterior
probability.  The mean empirical posterior of a state is the empirical
expectation $\delta[x^i]$ of the state divided by its frequency
$c(x^i)$.  The empirical expectation $\delta[x^i|y^T]$ of an \ith\
order state $x^i$ in an order $n$ mixture Markov model with respect to
a string $y^T$ is computed as follows
\begin{displaymath}
 \delta[x^i|y^T] = 
	\sum_{\{t : x^i = y_{t+1-i}^t\}} \delta(o_t=i|y^t)
,
\end{displaymath}
with the empirical posterior 
\begin{displaymath}
  \delta(o=i|y^t) = \frac{\delta(y_{t+1-i}^t)}{\sum_{j=1}^n \delta(y_{t+1-j}^t)}
.
\end{displaymath}
Note that $\delta[x^i|y^T]$ may be calculated for all states in
$O(nT)$ time using dynamic programming.  The empirical posterior
$\delta(o=i|y^t)$ of the \ith\ order state at time $t$ could be
weighted also by its predictive success $-\log
\delta(y_{t+1}|y_{t+1-i}^t)$.  A further refinement is to compute the
mean empirical posterior on withheld data.

As a final step, these values must be quantized to a finite number of
levels to construct the parameter tying scheme.

 \section{Conclusion}\label{conclude-section}

In this report, we propose a time series model that combines Markovian
events of varying orders using stochastic non-emitting transitions.
We prove that the resulting class of non-emitting Markov models is
strictly more powerful than the class of Markov models, including
interpolated and backoff models.  More importantly, our empirical
investigation reveals that the non-emitting model consistently
outperforms the strongest interpolated Markov models on natural
language texts, with only a modest increase in computational
requirements.

The expressive power of the non-emitting model comes from its ability
to represent additional information in its state order distribution.
To prove that the non-emitting model was strictly more powerful than
any Markov model, we used the state order distribution to represent an
unbounded dependency.  In our posthoc analysis, we revealed how the
model uses its hidden state order distribution to remember the
short-term effectiveness of all available Markovian states.

The non-emitting model succeeds empirically because it imposes a
pseudo-Bayesian discipline on maximum likelihood techniques.  The
interpolated model will favor a high-order state if it provides strong
predictions on withheld data.  The non-emitting model will favor a
high-order state if the state provides strong predictions on withheld
data {\it and it is well-predicted by the model\/}.  In order to reach
a high order state, the non-emitting model must assign high
probability to each symbol in that state.  Otherwise, the non-emitting
model will be forced to transition to a lower order state at a
previous time step and will not be able to reach the high order state.
Thus, the state occupancies of the non-emitting model are influenced
as much by their prior probabilities (pseudo-Bayes) as their past
ability to predict the future (maximum likelihood).

Finally, we note the use of non-emitting transitions is a general
modeling technique that may be employed in any time series model, for
symbolic domains and for continuous domains.

\clearpage
\appendix

 \section{Backoff}\label{backoff-appendix}

The backoff model is arguably the most widely used statistical
language model, due in large part to its ease of implementation,
computational efficiency, reasonable performance at lower model
orders, and an influential paper \cite{katz:87}.  Backoff models are
also widely used in the data compression community, in large part due
to their computational efficiency \cite{cleary-witten:84}.  Here we
review the backoff model, establish the equivalence of backoff models
and basic Markov models, and then specify a class of non-emitting
backoff models that is strictly more powerful than the class of
traditional backoff models.

In a backoff model, event probabilities are combined according to a
partial order.  Typically, higher order events are preferred over
lower order events.  The event probabilities are rescaled as we move
through the partial order so that the derived probability function is
valid.  The efficacy of the backoff model depends on the events that
are included in the model, their individual probabilities, and the
order in which they are combined.

Formally, a hierarchical backoff model $\theta = \tuple<A,E,\delta>$
consists of an alphabet $A$, a dictionary $E$ of selected state
transitions, $E\subseteq A^*\times A$, and the state transition
probabilities $\delta:E\rightarrow [0,1]$.  The state transition
probabilities $\delta$ are extended to an unbounded domain by
selecting the maximal suffix of the history that appears with the
relevant symbol in the dictionary $E$ of state transitions.
\begin{equation}\label{backoff-evaluate}
 p_b(y|x^t,\theta) = 
  \left\{\begin{array}{ll}
   \delta(y|x^t)			& \mbox{ if } \tuple<x^t,y>\in E \\
   \eta(x^t) p_b(y|x_2^t,\theta)	& \mbox{ otherwise }
  \end{array}\right.
\end{equation}
where $\eta(x^t)$ rescales the conditional probability distribution
as we backoff from higher order events to lower order events
\begin{displaymath}
  \eta(x^i) \doteq (1-\delta(E(x^i)|x^i)) / (1-p_b(E(x^i)|x_2^i))
\end{displaymath}
and $E(x^i)$ is the set of symbols available in the context $x^i$.
\begin{displaymath}
  E(x^i)\doteq \{ y : x^iy\in E \}
\end{displaymath}
The rescalar $\eta(x^i)$ is computed directly form the transition
probabilities $\delta(\cdot|x^i)$ in conjunction with the transition
dictionary $E$.  It is not a free parameter.

A hierarchical backoff model is a valid probability model if the
dictionary $E$ includes every $0^{\rm th}$ order state transition --
$\{\epsilon\}\times A\subset E$ -- and every $\delta(E(x^i)|x^i)$ is a
valid probability function.  A backoff model is nonzero for all
strings $A^*$ if every $\delta(y|x^i)$ is nonzero and no
$\delta(E(x^i)|x^i)$ is unity when $E(x^i) \subset A$.

In order to induce a hierarchical backoff model from data, we must
select the state transition dictionary and estimate its probabilities.
One simple -- but highly effective -- selection technique is to
include every state transition whose frequency exceeds a fixed
threshold, that may depend on the state order.  More effective
selection techniques require significant computational resources
\cite{ristad-thomas:acl95}.  The state transition probabilities
$\delta(y|x^t)$ are typically assigned by multinomial estimates,
either as conditional events $y|x^i$ in the symbol alphabet $A$ or as
joint events $x^iy$ in the string alphabet $A^{i+1}$.  The most widely
used multinomial estimates for statistical language modeling employ
some form of discounting
\cite{cleary-witten:84,good:53,good:65,ney-etal:95,howard-vitter:92},
although other estimators have also been shown to be effective
\cite{moffat:90b,rissanen:83c,ristad:pu495-95}.

A valid backoff model $\theta$ whose event dictionary $E$ is a subset
of $A^{n+1}$ can be converted into an equivalent basic Markov model
$\phi^\prime$ of order $n$, simply by setting
$\delta_n^\prime(x_t|x_{t-n}^{t-1})$ equal to
$p_b(x_t|x_{t-n}^{t-1},\theta)$.  Every basic model is a backoff model
with a complete state transition dictionary.  Consequently, the class
of backoff models is extensionally equivalent to the class of basic
Markov models.

The hierarchical non-emitting backoff model $\theta =
\tuple<A,E,\delta>$ has the same parameterization as the traditional
backoff model.  Unlike the traditional model, the backoff from the
state $x^i$ to its maximal proper suffix $x_2^i$ is permanent in the
non-emitting backoff model.
\begin{equation}\label{nonemit-backoff-evaluate}
 p_\epsilon(y^j|x^i,\theta) = 
  \left\{\begin{array}{ll}
   \delta(y_1|x^i) p_\epsilon(y_2^j|x^iy_1,\theta) & \mbox{ if } \tuple<x^i,y_1>\in E \\
   \eta(x^i) p_b(y^j|x_2^i,\phi)		   & \mbox{ otherwise }
  \end{array}\right.
\end{equation}
The rescalar $\eta(x^i)$ is identical in both version of the backoff
model.  

The class of non-emitting backoff models is strictly more powerful
than the class of basic Markov models, by a similar argument as in
lemma~\ref{state1-lemma}.  Although the backoff model does not have
any mixing parameters, we may use the presence or absence of a state
transition $y|x^i$ in the dictionary $E$ to control the hidden state
order.  Conversely, every order $n$ backoff model can be converted
into an equivalent non-emitting backoff model with a complete state
transition dictionary $E = A^{n+1}$.  Therefore, the class of
non-emitting backoff models is strictly more powerful than the class
of simple backoff models.

\end{document}